\documentclass{article}

\usepackage[preprint]{corl_2026} %

\usepackage{booktabs}
\usepackage{amsmath}
\usepackage{graphicx}
\usepackage{multirow}
\usepackage{url}
\usepackage{wrapfig}
\usepackage{siunitx}
\usepackage{array}

\newcommand{\website}[0]{\url{https://faerber-lab.github.io/RCT/}}

\title{RCT: A Robot-Collected Touch--Vision--Language Dataset for Tactile Generalization}

\author{
  Jingbo He\textsuperscript{1} \quad
  Michael Färber\textsuperscript{1} \quad
  Roberto Calandra\textsuperscript{2}\\
  \textsuperscript{1}ScaDS.AI, TU Dresden \quad
  \textsuperscript{2}LASR Lab, TU Dresden\\
  \texttt{\{jingbo.he, michael.faerber\}@tu-dresden.de, rcalandra@lasr.org}\\[4pt]
  {\normalfont\small Project page: \url{https://faerber-lab.github.io/RCT/}\quad Code: \url{https://github.com/faerber-lab/RCT}}
}

\hypersetup{
  pdfencoding=auto,
  pdftitle={RCT: A Robot-Collected Touch-Vision-Language Dataset for Tactile Generalization},
  pdfauthor={Jingbo He, Michael Färber, Roberto Calandra},
  pdfsubject={},
}

\begin{document}
\maketitle

\begin{abstract}
For robots manipulating open-world objects, tactile representations must generalize to unseen
materials. We introduce RCT (Robotic Contact Tactile), a robot-collected
touch--vision--language dataset with $29{,}279$ tactile frames from full robot presses on
$122$ industrial reference materials in $7$ categories, recorded with three DIGIT sensors at
multiple contact positions. RCT preserves each press as a \emph{contact sequence}, enabling
held-out evaluation across materials, categories, sensors, contact positions, and contact
sequences. 
Frames from one press are strongly correlated: 
frame-random
splits can place near-duplicate observations of the same physical interaction in both training
and test. 
With the encoder held fixed, 
removing contact-sequence overlap reduces
tactile$\rightarrow$text Recall@1 by $17.7$\,pp. 
When materials are additionally held out at training time,
performance drops sharply, 
leaving held-out-material Recall@1 at $25.1 \pm 6.1\%$ averaged over three held-out draws. The
public TVL/HCT split shows the same structure: every test contact sequence appears in training,
and raw-pixel nearest neighbors recover the correct sequence in $98.3\%$ of cases. 
Uniformly sampling a press improves contrastive training, and RCT-trained embeddings improve category
probes on unseen materials. 
RCT makes contact-sequence-aware, held-out-material evaluation reproducible and exposes novel-material generalization as a central challenge for robotic tactile perception.
The RCT dataset is open-sourced at \website{}
\end{abstract}

\keywords{Touch Sensing, Multimodal Representation Learning, Foundational Models, Evaluation Protocols}

\section{Introduction}
\label{sec:intro}

\begin{figure}[t]
  \centering
  \includegraphics[width=1\linewidth]{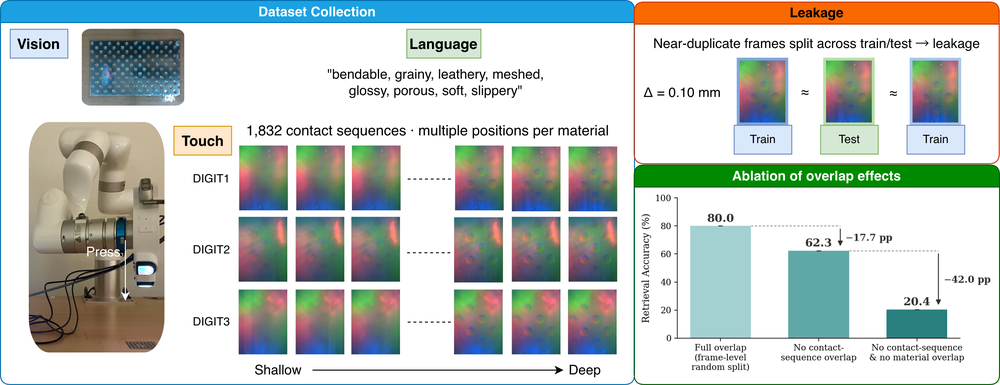}
  \caption{The \emph{RCT dataset} is the first dataset of its kind which preserves full contact sequences and enables held-out
  evaluation across materials, categories, sensors, contact positions, and robot presses.}
  \label{fig:teaser}
\end{figure}

Robots that manipulate objects in unstructured environments need to reason about the
materials they touch. A grasp that is safe for a rigid plastic part may fail on a soft foam,
a brittle composite, or a slippery textile. Vision reveals shape and appearance, but many
material properties that matter for manipulation -- such as compliance, texture,
deformability, and contact stability -- are observed most directly through touch. A useful
tactile representation must therefore do more than recognize contacts seen during training:
It must generalize to materials and contact conditions that the robot has not encountered
before.
Recent touch-vision-language learning has made important progress toward this goal. By
aligning vision-based tactile observations from sensors such as DIGIT~\citep{lambeta2020digit}
with object appearance and natural language, recent work~\citep{fu2024tvl, cheng2024touch100k,
yang2024unitouch, feng2025anytouch} has produced multimodal embedding spaces for tactile
retrieval, open-vocabulary recognition, and touch-conditioned description. These datasets and
models have broadened tactile representation learning, but they primarily emphasize scale,
multimodal alignment, or cross-sensor transfer. Less well isolated is a basic robotics question:
When a robot touches a previously unseen material, does the tactile representation generalize
beyond the specific contacts observed during training?

We introduce RCT, a robot-collected touch--vision--language dataset designed to study this
question. RCT records full robot presses on $122$ industrial reference materials from $7$
categories using three DIGIT sensors and multiple contact positions. Each material is paired
with visual and language annotations, and each tactile observation is stored with metadata for
controlled held-out evaluation. The key design choice is that RCT preserves the full
\emph{contact sequence}: the ordered tactile frames recorded during one robot press on one
material, with one sensor, at one contact position. This enables held-out evaluation across
materials, categories, sensors, contact positions, and contact sequences. To our knowledge, RCT is
the first robot-collected touch--vision--language dataset built explicitly around contact
sequences and controlled held-out evaluation of tactile generalization.
This structure matters because frames within one robot press are not independent samples. As the
sensor indents the material, successive tactile frames differ by only a small depth step
(\SI{0.10}{\milli\meter} in our setup) and are near-duplicate observations of the same physical interaction.
A frame-random split can therefore place frames from the same contact sequence in both training
and test. A model may then retrieve a near-duplicate contact observation rather than learn a
representation that transfers to new materials or contact conditions.

Using RCT, we show that this distinction has a large empirical effect. With the encoder held
fixed, removing contact-sequence overlap reduces retrieval by $17.7$ percentage points. When test
materials are also held out, performance drops sharply, and
held-out-material retrieval averages $25.1 \pm 6.1\%$ over three held-out draws. As an external check, we audit the released
TVL/HCT split and find the same structure: every test contact sequence also appears in training,
and a training-free nearest-neighbor baseline on raw tactile pixels recovers the correct sequence
in $98.3\%$ of cases. This does not invalidate touch--vision--language learning; it shows that
frame-random retrieval scores should not be interpreted as tactile generalization unless
contact-sequence and material overlap are controlled.
RCT also yields a practical training insight. Dense sampling of every frame in a press adds many
near-duplicate observations to contrastive learning. Uniformly sampling a small number of frames
from shallow to deep contact outperforms using all frames while requiring only a third of the
data, indicating that the contact sequence is a useful unit for both evaluation and training.

\paragraph{Contributions.}
This paper contributes: \textbf{(1)} RCT, a robot-collected touch--vision--language dataset with
full contact sequences, material-level vision and language annotations, and metadata for
controlled held-out evaluation; \textbf{(2)} protocols for evaluating generalization across
materials, categories, sensors, contact positions, and robot presses; \textbf{(3)} empirical
evidence that frame-random splits overestimate tactile generalization by mixing contact-sequence
overlap and material overlap; and \textbf{(4)}~a contact-sequence sampling result showing that
uniformly sampling a press improves contrastive training. Together, these results identify
novel-material generalization as a central open challenge for robotic tactile perception. 

\section{Related Work}
\label{sec:related}

\paragraph{Tactile datasets and representation learning.}
Optical tactile sensing, introduced by GelSight~\citep{yuan2017gelsight} and miniaturized in
DIGIT~\citep{lambeta2020digit}, made dense visuo-tactile data collection practical, with early work
linking such readings to grasp outcomes~\citep{calandra2017feeling}. Visuo-tactile datasets range
from human-collected in-the-wild touch~\citep{yang2022touchandgo}, simulated multisensory
objects~\citep{gao2022objectfolder}, and tactile simulators~\citep{wang2020tacto} to task-oriented
visuotactile pairs~\citep{kerr2023ssvtp} and touch--vision--language triplets, which scale to over
100k entries~\citep{cheng2024touch100k} or emphasize finer-grained sentence-level language
supervision~\citep{cheng2024tlv}; large tactile--language models further use such data for
object-property reasoning~\citep{yu2024octopi}. Beyond these, general-purpose tactile encoders are
learned across sensors or modalities~\citep{higuera2024sparsh, yang2024unitouch, feng2025anytouch}.
A long line of work also uses vision-based tactile sensing to recognize material and surface
properties directly~\citep{yuan2017lookfeel, yuan2018activecloth}, but typically evaluates on
materials seen during training; RCT instead targets generalization to held-out materials. These
efforts emphasize scale, sensor coverage, multimodal alignment, or cross-sensor transfer, and recent
work also reduces collection cost through robot-free, human-held
acquisition~\citep{wu2026freetacman}. RCT is complementary: it uses a controlled robot-press
protocol organized around contact sequences, with controlled held-out evaluation across materials,
categories, positions, and sensors.

\paragraph{Multimodal alignment.}
Contrastive vision--language pretraining~\citep{radford2021clip, jia2021align, zhai2023siglip}
and unified embedding spaces~\citep{girdhar2023imagebind} motivate aligning touch into a shared
semantic space; TVL applies InfoNCE-style alignment across tactile, vision, and
text~\citep{fu2024tvl}. We use this recipe unchanged and focus on evaluation.

\paragraph{Evaluation, distribution shift, and leakage.}
Near-duplicate train/test overlap is known to inflate performance in vision
benchmarks~\citep{barz2020cifair}, and more broadly is a recurring source of leakage documented
across ML-based science~\citep{kapoor2023leakage}. A parallel line of work argues that models should
be evaluated under deployment-relevant distribution shift rather than i.i.d.\ test splits, as
formalized in benchmarks such as WILDS~\citep{koh2021wilds}. In tactile sensing specifically,
cross-sensor domain gaps have motivated dedicated domain-adaptation
methods~\citep{jing2025crosssensor}, yet evaluation typically still allows train/test sharing of the
same materials and physical interactions. Tactile data admits a more structural case of leakage: one
robot press produces many highly correlated frames. We are not aware of prior work that quantifies
contact-sequence overlap in tactile representation-learning benchmarks. RCT makes the contact
sequence and the material the held-out unit, connecting tactile evaluation to the distribution-shift
framing above.

\section{The RCT Dataset}
\label{sec:dataset}

RCT is a robot-collected touch-vision-language dataset for evaluating tactile generalization
under controlled held-out conditions. Its central design choice is to store tactile data as
\emph{contact sequences}: the ordered frames recorded during one robot press on one material,
with one sensor, at one contact position. This structure enables held-out evaluation across
materials, categories, sensor instances, contact positions, and contact sequences.

\vspace{-0.1cm}
\textbf{Collection protocol.}
RCT is collected with a robot arm carrying a rotating adapter that holds three
DIGIT~\citep{lambeta2020digit} vision-based tactile sensors. Rotating the adapter brings one
sensor into the contact pose, allowing each material to be recorded with all three sensors under
the same protocol. For each material, the robot presses the surface at multiple discrete contact
positions. Each press is sampled from initial contact to deeper indentation at fixed \SI{0.10}{\milli\meter}
steps, typically yielding $15$--$17$ tactile frames per contact sequence; contact force is
recorded for each frame.

\begin{figure}[t]
  \centering
  \includegraphics[width=0.8\linewidth]{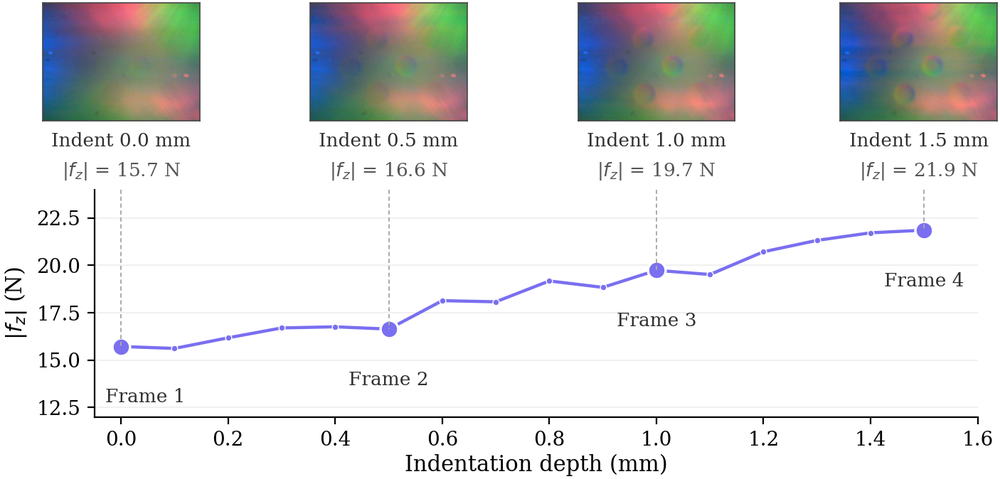}
  \caption{Example of contact sequence from one robot press. DIGIT frames evolve gradually with
  indentation depth, while force increases over the same press. Adjacent frames are correlated
  observations of one physical interaction, motivating contact-sequence-aware evaluation.}
  \label{fig:contact_sequence}
\end{figure}

\textbf{Materials and scale.}
RCT contains $122$ industrial reference materials from the Musterkiste material sample set
(Modulor GmbH), spanning $7$ top-level categories: Plastic/Rubber ($33$), Paper/Cardboard ($45$),
Metal ($20$), Textiles/Leather ($10$), Wood/Bamboo/Cork ($7$), Crafts ($4$), and Small Items
($3$). In total, RCT contains $29{,}279$ tactile frames, each indexed by material, contact
position, sensor, contact sequence, and indentation depth. Contact positions $1$--$4$ are densely
sampled across all materials, while later positions are sparser; we report this imbalance as metadata and discuss its
implications in Sec.~\ref{sec:limitations}.

\begin{table}[t]
  \centering
  \footnotesize
  \caption{Overview of RCT. The dataset is organized around contact sequences.} %
  \label{tab:rct_overview}
  \begin{tabular}{ll}
    \toprule
    Property & RCT \\
    \midrule
    Collection & Robot-collected presses \\
    Sensor & Three DIGIT vision-based tactile sensors \\
    Materials / categories & $122$ materials / $7$ categories \\
    Tactile frames & $29{,}279$ \\
    Structural unit & Contact sequence, i.e., one robot press \\
    Contact variation & Multiple positions and indentation depths \\
    Additional signal & Per-frame contact force \\
    Modalities & Touch, material image, language descriptors \\
    Supported holdouts & Contact sequence, material, category, sensor, position \\
    Release & Dataset and split-generation toolkit \\
    \bottomrule
  \end{tabular}
\end{table}

\textbf{Vision and language modalities.}
Each material is paired with one high-resolution studio photograph ($2048\times1536$) and
material-level perceptual descriptors. This separates \emph{material identity} from
\emph{contact state}: the image and descriptors describe the material, while tactile frames
capture how contact evolves during a press. Since many tactile frames share the same material
image, strict sample-level tactile$\rightarrow$vision retrieval is unsuitable; we use a
material-level multi-positive criterion instead. For language, two human annotators independently
touched all $122$ materials and selected descriptors from a fixed vocabulary; we also include a
vision-language model source~\citep{gemma2025gemma3}. The default target is the
union of the two human annotations. Human agreement is moderate (Jaccard $0.294$), while
VLM--human agreement is lower ($0.110$ and $0.108$), indicating that visual descriptors capture
only part of what human annotators report from touch.

\section{Evaluation Protocol}
\label{sec:protocol}

Our evaluation asks whether tactile representations generalize across physical variation that
matters in robot deployment. The central unit is the \emph{contact sequence}: all frames recorded
during one robot press on one material, with one sensor, at one contact position. Frames within a
sequence are strongly correlated; in RCT, successive frames differ by only a \SI{0.10}{\milli\meter}
indentation step. A frame-random split can therefore place near-duplicate observations of the
same physical interaction in both training and test. Our protocols instead hold out physically
meaningful units.

\textbf{Held-out settings.}
Unless stated otherwise, we use held-out-material evaluation with $K=20$ materials as the primary
setting, since it matches the deployment question: can a robot recognize material properties for
materials unseen during training? RCT also supports held-out-category, held-out-contact-sequence,
held-out-contact-position, and held-out-sensor evaluation. The frame-random setting is used only
as a control to quantify the effect of splitting individual frames while allowing material and
contact-sequence overlap. Table~\ref{tab:splits} summarizes the settings.

\begin{table}[t]
  \centering
  \footnotesize
  \setlength{\tabcolsep}{4pt}
  \caption{Held-out evaluation settings supported by RCT.}
  \label{tab:splits}
  \begin{tabular}{@{}>{\raggedright\arraybackslash}p{0.21\linewidth}>{\raggedright\arraybackslash}p{0.49\linewidth}>{\raggedright\arraybackslash}p{0.22\linewidth}@{}}
    \toprule
    Setting & Held out at test time & Purpose \\
    \midrule
    Held-out material & All contact sequences of $K$ materials ($K=2,5,20$) & Unseen materials \\
    Held-out category & All materials from one top-level category & Unseen material family \\
    Held-out sequence & Entire contact sequences; materials may be seen & Sequence overlap \\
    Held-out position & One contact position across materials & Novel location \\
    Held-out sensor & One DIGIT sensor instance & Sensor transfer \\
    Frame-random control & Random frames; material and sequence overlapped & Split inflation \\
    \bottomrule
  \end{tabular}
\end{table}

\textbf{Sampling, model, and metrics.}
Independently of the split, we vary how many frames represent a contact sequence: \textbf{full}
uses all depths ($\sim$16 frames), \textbf{uniform5} uses five frames evenly spaced from shallow
to deep contact, and \textbf{deep5} uses the five deepest frames. To isolate the effect of the
evaluation protocol, we follow the TVL training recipe~\citep{fu2024tvl}: a ViT-Base tactile
encoder, OpenCLIP ViT-L/14 vision and text encoders, batch size $256$, $200$ epochs, a
$10$-epoch warmup, base learning rate $1.5\!\times\!10^{-4}$, weight decay $0.05$, cosine
schedule, and InfoNCE losses across modality pairs. We report Recall@1 and Recall@5. For
RCT tactile$\rightarrow$vision, we use the material-level multi-positive criterion from
Sec.~\ref{sec:dataset}; for tactile$\rightarrow$text, we follow TVL's soft-label criterion based
on text--text cosine similarity with threshold $0.6356$.

\section{Experimental Evaluation}
\label{sec:experiments}

The experiments test whether tactile representations generalize to new materials and contact
conditions, or instead benefit from frame-level overlap. We first verify that our implementation
reproduces TVL, then use RCT to separate contact-sequence overlap from material overlap, audit the
released TVL/HCT split for the same structure, and test whether RCT improves training and
material-level recognition.

\textbf{Sanity check: reproducing TVL.}
Using the published TVL recipe \citep{fu2024tvl} on 
SSVTP+HCT, our model reaches $82.21\%$ tactile$\rightarrow$vision Recall@1 
on the combined 402-item retrieval pool, close to the $81.7\%$ 
reported in TVL~\citep{fu2024tvl}. Table~\ref{tab:repro} reports the full 
reproduction.

\begin{table}[t]
  \centering
  \footnotesize
  \caption{TVL reproduction. Tactile$\rightarrow$text uses the TVL soft-label criterion with
  threshold $0.6356$.}
  \label{tab:repro}
  \begin{tabular}{lccc}
    \toprule
    Evaluation pool & Test samples & Tac$\rightarrow$vis R@1 & Tac$\rightarrow$text R@1 \\
    \midrule
    SSVTP test           & 46  & 36.96\% & 33.70\% \\
    HCT test             & 356 & 88.06\% & 34.83\% \\
    Combined (SSVTP+HCT) & 402 & 82.21\% & 34.95\% \\
    \bottomrule
  \end{tabular}
\end{table}

\begin{wrapfigure}{r}{0.58\textwidth}
  \centering
  \includegraphics[width=\linewidth]{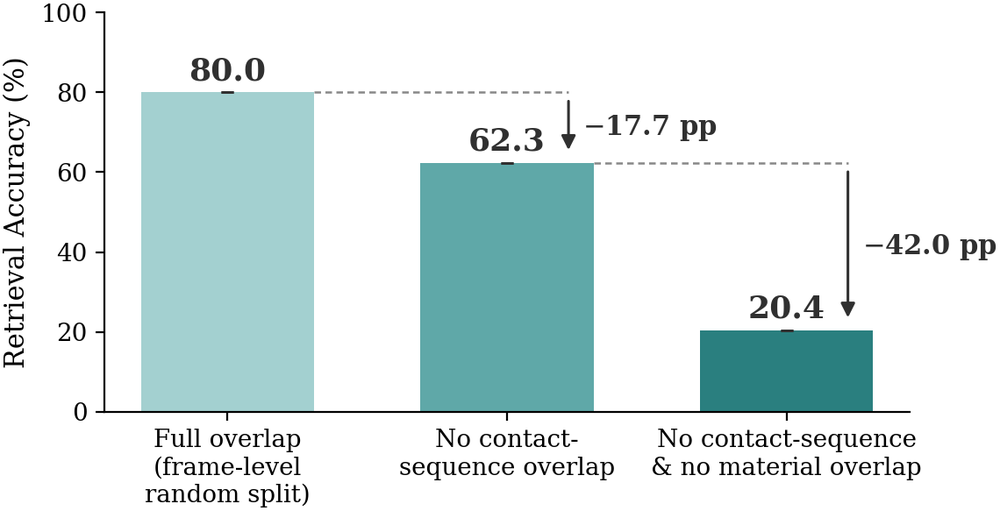}
  \caption{Decomposing frame-random performance. Removing contact-sequence overlap costs
  $17.7$\,pp; additionally holding out materials costs a further $42.0$\,pp. Values are for one
  representative held-out split (seed 42).}
  \label{fig:staircase}
\end{wrapfigure}

\textbf{Contact-sequence overlap inflates frame-random evaluation.}
\label{sec:leakage}
We evaluate tactile$\rightarrow$text Recall@1 under three conditions (Figure~\ref{fig:staircase}):
a frame-random control, a held-out-contact-sequence setting where materials may still be seen in
training, and a held-out-material setting where all test contact sequences and materials are
unseen.
The first drop isolates contact-sequence overlap: the encoder is unchanged, but the test pool no
longer contains contact sequences seen during training. Recall@1 falls from $80.01\%$ to
$62.33\%$. An encoder trained directly under the held-out-contact-sequence protocol reaches
$62.44\%$ on the same pool, confirming that the effect is not due to a weaker representation.
The second drop, from $62.33\%$ to $20.35\%$, is the material-overlap effect: contacts from
materials seen during training are far easier than contacts from materials never seen before.

\textbf{Adjacent frames explain part 
of the effect.}
Adjacent near-duplicate frames are one mechanism behind contact-sequence overlap. When we
uniformly subsample each press at test time, the frame-random gain decreases from $17.7$ to
$12.3$ percentage points; keeping only the five deepest frames decreases it to $10.4$ percentage
points. Thus, removing immediate neighbors reduces the effect, but a substantial gap remains,
suggesting that models also exploit broader sequence-specific structure, such as material-,
sensor-, and position-specific contact patterns that persist across the press.

\textbf{The released TVL/HCT split has the same structure.}
\label{sec:public}
Using the released split files, we define an HCT contact sequence by its contact-session folder
(\texttt{<idx>-<timestamp>}) and restrict to in-contact frames. All $279$ HCT test contact
sequences also appear in training (Table~\ref{tab:audit}). The median nearest same-sequence
training frame is one step away from the test frame.
\begin{table}[t]
  \centering
  \footnotesize
  \caption{Contact-sequence overlap in the released HCT/TVL split. SSVTP contains one frame per
  contact and therefore has no sequence structure.}
  \label{tab:audit}
  \begin{tabular}{lcccc}
    \toprule
    Dataset & Test frames & Test contact seq. & Test seq. in train & Median gap \\
    \midrule
    SSVTP & 46  & --- & --- & --- \\
    HCT   & 356 & 279 & 279 (100\%) & 1 step \\
    TVL   & 402 & 279 (HCT only) & 279 (100\%) & 1 step \\
    \bottomrule
  \end{tabular}
\end{table}
This overlap is exploitable without learning. For each HCT test frame, we compute raw-pixel L2
distance to the nearest same-sequence training frame and nearest cross-sequence frame. In
$350/356$ cases ($98.3\%$), the same-sequence neighbor is closer, with median cross/same ratio
$2.6\times$. A training-free raw-pixel nearest-neighbor baseline therefore recovers the correct
contact sequence with $98.3\%$ top-1 accuracy. This does not invalidate TVL as a multimodal
tactile-learning framework; it shows that frame-random scores should not be interpreted as
tactile generalization unless contact-sequence overlap is controlled.

\textbf{Uniformly sampling a press improves training.}
\label{sec:recipe}
Dense use of all frames in a press is not always beneficial. We compare three training densities
and evaluate all models on the same held-out-contact-sequence test pool at full density
(Table~\ref{tab:recipe}).
\begin{table}[t]
  \centering
  \footnotesize
  \caption{Training-time frame sampling, evaluated on the same held-out-contact-sequence test
  pool.}
  \label{tab:recipe}
  \begin{tabular}{lcc}
    \toprule
    Training density & Training frames & Tac$\rightarrow$text R@1 \\
    \midrule
    Full press ($\sim$16 frames/sequence) & 22{,}576 & 62.44\% \\
    Deep5 (five deepest frames)           & 7{,}055  & 59.52\% \\
    Uniform5 (five frames across press)   & 7{,}055  & \textbf{68.18\%} \\
    \bottomrule
  \end{tabular}
\end{table}
Uniform5 improves over full-density training by $5.7$ percentage points while using only one
third of the frames, and over deep5 by $8.7$ points at equal data volume. The trend also holds for
tactile$\rightarrow$vision under the multi-positive criterion ($78.29\%$ vs. $70.95\%$ for full).
Thus, covering the contact sequence from shallow to deep contact is more useful than densely
sampling near-duplicate frames. This is consistent with evidence that near-duplicate items act as false negatives and can degrade contrastive objectives~\citep{chuang2020debiased}.

\textbf{Held-out-material generalization remains hard.}
\label{sec:novel}
In the deployment-relevant held-out-material setting, RCT-only training reaches $20.4\%$
tactile$\rightarrow$text Recall@1 for $K=20$ unseen materials (averaged over three held-out
draws, $25.1 \pm 6.1\%$; Table~\ref{tab:app_seed}). Category holdouts are similarly
difficult (Table~\ref{tab:novel}), and adding SSVTP+HCT to the training data does not improve
performance in our setup. These results identify novel-material generalization as the central
open challenge exposed by RCT.

\begin{table}[t]
  \centering
  \footnotesize
  \setlength{\tabcolsep}{4pt}
  \caption{Held-out-material and held-out-category generalization. Results are
  tactile$\rightarrow$text Recall@1 (\%) using the human-union annotations.}
  \label{tab:novel}
  \begin{tabular}{lccccc}
    \toprule
    Training data & Material $K{=}20$ & Cat-Metal & Cat-Plastic & Cat-Paper & Cat-Textiles \\
    \midrule
    RCT only          & \textbf{20.4} & \textbf{30.2} & \textbf{26.0} & \textbf{32.1} & \textbf{20.0} \\
    SSVTP+HCT+RCT     & 13.4 & 28.7 & 18.0 & 30.2 & 12.4 \\
    \bottomrule
  \end{tabular}
\end{table}

\begin{wraptable}{r}{0.6\textwidth}
\vspace{-1.5em}

  \centering
  \footnotesize
  \setlength{\tabcolsep}{4pt}
  \caption{Material structure and downstream probes. Separability margins are higher-is-better and are computed on the deepest-per-position frame of each held-out material.
  $K=2$ and $K=5$ use diverse selection (single encoder); $K=20$ uses category-balanced selection
  averaged over three RCT encoders and three held-out material sets. Category is a 7-class
  problem (majority $22\%$); hard/soft is binary (majority $59\%$).}
  \label{tab:embedding_structure}
  \begin{tabular}{lcccc}
    \toprule
    \multicolumn{5}{l}{\textbf{Material separability across held-out set sizes}} \\
    Encoder & $K{=}2$ & $K{=}5$ & $K{=}20$ & \\
    \midrule
    TVL encoder & 0.086 & 0.039 & $0.048 \pm 0.010$ & \\
    RCT encoder & \textbf{0.266} & \textbf{0.215} & $\mathbf{0.297 \pm 0.020}$ & \\
    Random init & 0.011 & 0.005 & $0.004 \pm 0.001$ & \\
    \midrule
    \multicolumn{5}{l}{\textbf{Embedding structure and downstream probes on $K=20$}} \\
    Metric & TVL & RCT & Random & Force \\
    \midrule
    Separability margin    & 0.048 & \textbf{0.297} & 0.004 & --- \\
    Category acc. (\%)     & 34.2  & \textbf{49.8}  & 37.8  & --- \\
    Hard/soft acc. (\%)    & 52.2  & 52.2           & 45.7  & 52.6 \\
    \bottomrule
  \end{tabular}
\end{wraptable}
\textbf{RCT improves material structure in the embedding space.}
\label{sec:downstream}
Finally, we evaluate whether RCT training improves material-level structure on held-out
materials. We measure a material-separability margin
$m=d_{\mathrm{inter}}-d_{\mathrm{intra}}$, where $d_{\mathrm{intra}}$ is the mean cosine distance
among frames of the same material and $d_{\mathrm{inter}}$ the mean distance to other materials.
The RCT encoder has a larger margin than the released TVL encoder across held-out set sizes
(Table~\ref{tab:embedding_structure}), indicating that the effect is not restricted to a particular
test set. We then train linear probes on frozen per-sequence embeddings for material category and
binary hard/soft prediction, using the same table to relate geometric structure to downstream
predictability.
The separability gain translates into stronger downstream category recognition: a linear probe on
RCT embeddings reaches $49.8\%$ on unseen materials, compared with $34.2\%$ for TVL and
$37.8\%$ for a random-initialized encoder. Hard/soft prediction remains below the majority
baseline for all learned representations and for force features, suggesting that category-level
material information transfers better than binary hardness in the current setup.

\section{Discussion: Benchmarking Tactile Generalization}
\label{sec:discussion}

RCT is intended to make tactile representation results easier to interpret, not to replace every
existing evaluation setting. Frame-random retrieval is still useful as an engineering diagnostic:
it verifies that a model can align modalities and recognize contacts drawn from the same data
collection distribution. What it does not show is whether the representation transfers to a new
physical interaction, material, sensor, or contact location. Our results suggest that tactile
benchmarks should therefore report the level at which test samples are independent from training.
For vision-based tactile sensors, this level is at least the contact sequence, and for robotic
manipulation it is often the material or object being touched.

\textbf{Reporting recommendations.}
The practical implication is a small change in benchmark design: tactile datasets should expose
contact-sequence identifiers and report held-out results alongside any frame-random scores. This
is especially important for contrastive touch--vision--language training, where many positives and
negatives are generated from repeated observations of the same surface. Without contact-sequence
metadata, a high retrieval score may reflect memorization of a press rather than material-level
understanding. %

\textbf{What the current results do and do not imply.}
The TVL/HCT audit should be read as an evaluation result, not as evidence that touch--vision--language
alignment is unhelpful. In fact, RCT uses the same training recipe and shows that tactile embeddings
can improve material structure and category recognition. The issue is attribution: if the test set
contains frames from contact sequences seen during training, then retrieval accuracy cannot be
interpreted as evidence of generalization to new materials or new interactions. Conversely, low
held-out-material performance should not be read as failure of the dataset; it identifies the regime
where the field still lacks robust methods.

\textbf{Directions enabled by RCT.}
RCT makes several method directions testable. First, models can use the full contact sequence rather
than treating frames as independent images, for example by pooling over depth, modeling temporal or
indentation order, or learning depth-aware contrastive objectives. Second, sampling can be optimized:
our uniform5 result suggests that coverage across the press is more valuable than simply using more
near-duplicate frames. Third, held-out sensor, position, and category splits make it possible to ask
which invariances a tactile encoder learns and which remain brittle. These questions are central for
robots that must use touch outside the exact contacts observed during data collection.

\section{Limitations and Failure Modes}
\label{sec:limitations}

\textbf{Scope and supervision.}
RCT is designed for controlled evaluation, not exhaustive open-world tactile exploration. It uses
$122$ industrial reference materials, three DIGIT sensors, and a robot-press protocol that enables
held-out evaluation across materials, categories, sensors, contact positions, and contact
sequences, but it does not yet cover arbitrary object geometries, curved surfaces, or dynamic
exploratory motions. RCT also uses material-level supervision: each material has one reference
image and material-level descriptors, not pose-conditioned per-frame visual targets or
frame-specific tactile captions. Thus, strict one-to-one tactile$\rightarrow$vision retrieval is
not appropriate; material-level multi-positive evaluation is required. 

\textbf{Failure modes and curation.}
RCT improves material separability and category recognition on unseen materials, but binary
hard/soft prediction remains close to the majority baseline, even with summary force--depth
features. Hardness is a more subtle property than category: human labels carry subjective variation, and scalar summary features through a linear probe may not capture the curve-shape cues most informative of hardness on unseen materials. This suggests that some physical properties require richer interaction signals, more
targeted labels, or different probing tasks. Our TVL/HCT audit is an evaluation-protocol result,
not a criticism of touch--vision--language learning as a framework: frame-random scores should not
be interpreted as tactile generalization unless contact-sequence overlap is controlled. %

\section{Conclusion}
\label{sec:conclusion}

We introduced RCT, a robot-collected touch--vision--language dataset organized around full contact
sequences and controlled held-out evaluation. The main lesson is that the independent unit in
tactile evaluation is the robot press, not the individual frame: frame-random splits can place
near-duplicate observations of the same physical interaction in both training and test. Using RCT,
we show that removing contact-sequence overlap reduces retrieval by $17.7$ percentage points, and holding out materials reduces it by a further $42.0$ points on this split. The released TVL/HCT split exhibits
the same overlap structure, showing that this is not only an artifact of our dataset.
RCT also shows how tactile data should be used: uniformly sampling frames across a press improves
contrastive training, and RCT-trained embeddings improve material separability and category
recognition on unseen materials. Yet held-out-material retrieval averages $25.1 \pm 6.1\%$ over three held-out draws, and binary
hardness remains difficult. We therefore recommend contact-sequence-aware and held-out-material
evaluation as default reporting, and identify generalization to unseen materials and contact
conditions as a central challenge for robotic tactile perception. By making novel-material generalization measurable, we hope RCT helps the community develop tactile representations that robots can trust on unfamiliar materials, a prerequisite for safe and general-purpose manipulation in the real world.

\clearpage
%\acknowledgments{To be added in the camera-ready.}

\bibliography{references}

\clearpage
\appendix

\section{Additional Dataset and Annotation Details}
\label{app:dataset}

\begin{figure}[h]
  \centering
  \includegraphics[width=0.32\linewidth]{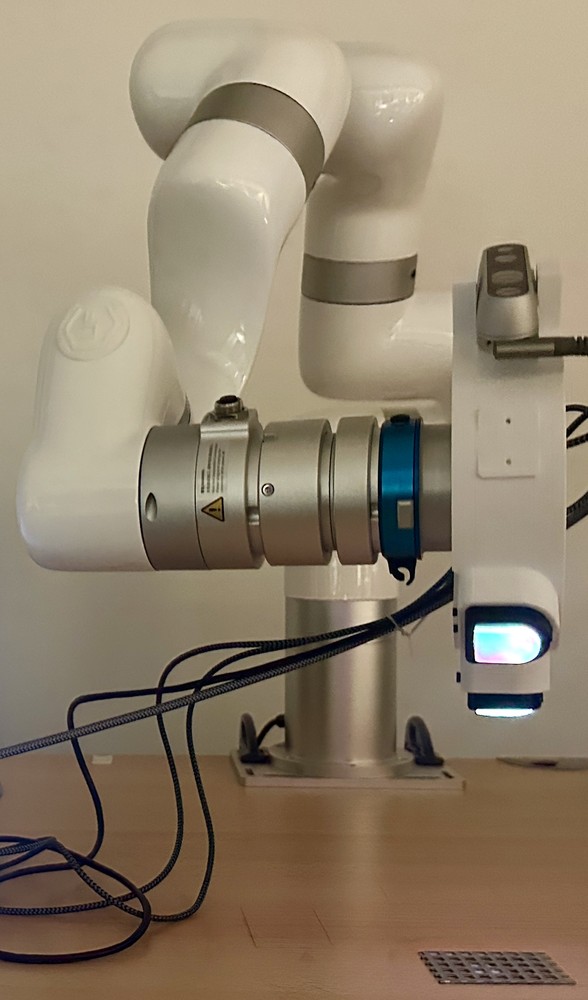}
  \caption{RCT collection setup. A robot arm presses industrial reference materials with a
  rotating adapter holding three DIGIT vision-based tactile sensors.}
  \label{fig:app_setup}
\end{figure}

\textbf{Contact-sequence metadata.}
Each tactile frame is indexed by material, material category, sensor instance, contact position,
contact sequence, and indentation depth. The released metadata will include these identifiers,
force signals, material-image identifiers, annotation-source identifiers, and all split assignments
used in the paper.

\textbf{Contact-position distribution.}
The contact-position distribution is heavy-tailed. Positions $1$--$4$ are densely sampled
across all materials ($\sim$5,800 frames each), position 5 covers 75 of 122 materials ($\sim$3,500 frames), and positions $6$--$18$ are progressively sparser, down to
roughly $50$ frames. This is why held-out-contact-position evaluation is treated as a diagnostic
for robustness to contact location rather than a complete benchmark for spatial generalization.

\textbf{Material-level visual supervision.}
RCT pairs each material with one high-resolution studio photograph. This creates a one-to-many
relationship between vision and touch: many tactile frames, recorded at different contact
positions, depths, and sensors, correspond to the same material image. For tactile$\rightarrow$vision
retrieval, we therefore use a material-level multi-positive criterion: all tactile samples linked
to the same material image are valid positives. A strict sample-level diagonal criterion would
incorrectly treat other tactile frames from the same material as negatives.

\textbf{Language annotation sources.}
Each material is annotated with material-level perceptual descriptors. Two human annotators
physically touched all materials and selected terms from a fixed vocabulary. We also include a
vision-language model (VLM) source that selects descriptors from the material photograph, and a
human-union source used as the default language target in the main experiments.
Table~\ref{tab:app_annotation_agreement} summarizes agreement between sources.

\begin{table}[h]
  \centering
  \footnotesize
  \caption{Agreement between descriptor sources, measured as Jaccard overlap over selected
  descriptor sets. Human--VLM agreement is substantially lower than human--human agreement,
  supporting the use of the human-union descriptors as the default tactile-language target.}
  \label{tab:app_annotation_agreement}
  \begin{tabular}{lc}
    \toprule
    Comparison & Jaccard overlap \\
    \midrule
    Human annotator 1 vs. human annotator 2 & $0.294$ \\
    VLM vs. human annotator 1 & $0.110$ \\
    VLM vs. human annotator 2 & $0.108$ \\
    Materials with no VLM--human overlap & $38\%$ \\
    \bottomrule
  \end{tabular}
\end{table}

The VLM tends to select visually apparent descriptors such as ``smooth'', ``cool'', ``hard'',
``flat'', and ``uniform'', whereas human annotators more often select touch-specific descriptors
such as ``elastic'', ``rubbery'', ``sharp'', and ``glassy''. This indicates that visual descriptors
capture only part of what humans report from touch.

\textbf{VLM source selection.}
The default language target is the union of the two human annotations; we additionally include a
VLM descriptor source for scale. To choose the VLM, we compared three open vision-language models, LLaVA-v1.6-Mistral-7B~\citep{liu2024llava}, Qwen2-VL-7B-Instruct~\citep{wang2024qwen2vl}, and Gemma-3-27B-IT~\citep{gemma2025gemma3}, under identical prompts along three axes: template dependence (how often a model reuses a fixed descriptor combination across
materials), lexical richness (unique descriptor tokens relative to total tokens), and semantic
alignment with the human annotations. Table~\ref{tab:app_vlm_selection} reports the comparison.
Gemma-3 is the least templated ($101$ distinct descriptor combinations across $122$ materials,
against $40$ for LLaVA and $23$ for Qwen2-VL), the most lexically rich ($11.15\%$, close to the
$10.81$--$11.21\%$ of the two human groups), and the closest to the human annotations ($26.36\%$
semantic similarity, about $59\%$ of the $44.43\%$ inter-annotator baseline). We therefore use
Gemma-3 as the VLM descriptor source. Even the best model stays well below the human-human
baseline, consistent with the main-paper finding that visual descriptors capture only part of
what human annotators report from touch. The full descriptor vocabulary ($54$ tactile terms) and
the VLM prompts are provided in the supplementary material.

\begin{table}[h]
  \centering
  \footnotesize
  \setlength{\tabcolsep}{5pt}
  \caption{VLM source selection. Three open vision-language models compared under identical
  prompts. Distinct combinations are out of $122$ materials; template dependence is
  lower-is-better and the other metrics are higher-is-better. The human reference row reports the
  two annotator groups (lexical richness) and their inter-annotator semantic-similarity baseline.
  Best VLM per column in bold.}
  \label{tab:app_vlm_selection}
  \begin{tabular}{lcccc}
    \toprule
    VLM & Distinct comb.\ ($/122$) & Template dep.\ (\%) & Lexical richness (\%) & Semantic sim.\ (\%) \\
    \midrule
    Qwen2-VL-7B   & $23$           & $81.15\%$          & $4.26\%$           & $16.58\%$ \\
    LLaVA-v1.6-7B & $40$           & $67.21\%$          & $6.07\%$           & $17.88\%$ \\
    Gemma-3-27B   & $\mathbf{101}$ & $\mathbf{17.21\%}$ & $\mathbf{11.15\%}$ & $\mathbf{26.36\%}$ \\
    \midrule
    Human reference & ---          & ---                & $10.81$--$11.21\%$ & $44.43\%$ \\
    \bottomrule
  \end{tabular}
\end{table}

\section{Annotation Details}
\label{app:annotation}

\subsection{Descriptor Vocabulary}
\label{app:vocab}

Human annotators selected 3--5 terms per material from the following fixed vocabulary of 54
tactile descriptors, grouped into five categories.

\textbf{Physical properties:} hard, soft, rigid, yielding, bendable, firm, elastic, brittle,
spongy, thick, thin.

\textbf{Surface texture:} cool, smooth, rough, silky, glossy, tarnished, bumpy, ridged, grooved,
corrugated, grainy, cottony, slippery, sharp, matte, transparent, translucent, fuzzy, scratchy,
fleecy, concave, convex.

\textbf{Material properties:} metallic, wooden, leathery, rubbery, papery, glassy, reflective,
honeycomb.

\textbf{Structure characteristics:} woven, knitted, crocheted, meshed, pleated, patterned,
fibrous, porous.

\textbf{Shape characteristics:} flat, carved, cushioned, rusty.

\subsection{VLM Prompt}
\label{app:prompt}

We used the following prompt to generate VLM descriptor pseudo-labels for the training set. Each
material was presented with two images: the full contact surface and a cropped view of the
material region.

\begin{verbatim}
Surface Type: [Specify the surface type, e.g., 'metal,' 'fabric']
Images: The first image shows the complete contact surface captured by camera,
including the material sample. The second image is a cropped version focusing
only on the material portion from the first image.
Example: For a smooth and cold surface, the description might be
'slick, chilly, hard, unyielding, glossy.'
Task: Based on these images, describe the possible tactile feelings of the
material surface using sensory adjectives. Limit your response up to five
adjectives, separated by commas.
\end{verbatim}

\section{Additional Protocol Details}
\label{app:protocol}

\textbf{Chance-normalized Recall@1.}
When candidate-pool sizes differ, we additionally compute chance-normalized Recall@1,
\begin{equation}
\mathrm{cR@1} = \frac{\mathrm{Recall@1}}{1/N} = N \cdot \mathrm{Recall@1},
\label{eq:app_chance_norm}
\end{equation}
where $N$ is the candidate-pool size. This measures how many times above chance a model retrieves
the correct target. Raw Recall@1 and Recall@5 remain the primary metrics in the main paper.

\textbf{Held-out material selection.}
For held-out-material experiments, test materials are selected greedily to maximize diversity over
the material-level descriptor vocabulary. This avoids test sets dominated by near-synonymous
materials and increases perceptual coverage of the held-out pool. Category holdouts use the
Musterkiste catalog hierarchy directly.

\textbf{Frame-density settings.}
We use three frame-density settings. \emph{Full} uses all depths in a contact sequence
($\sim$16 frames). \emph{Uniform5} uses five frames evenly spaced from shallow to deep contact.
\emph{Deep5} uses the five deepest frames only. These settings are used both as test-time controls
for contact-sequence overlap and as training-time sampling choices.

\textbf{Training configuration.}
Table~\ref{tab:app_training} lists the training configuration used throughout the paper.

\begin{table}[h]
  \centering
  \footnotesize
  \caption{Training configuration.}
  \label{tab:app_training}
  \begin{tabular}{ll}
    \toprule
    Component & Setting \\
    \midrule
    Tactile encoder & ViT-Base (\texttt{vit\_base\_patch16\_224}) \\
    Vision encoder & OpenCLIP ViT-L/14 \\
    Text encoder & OpenCLIP ViT-L/14 \\
    Loss & InfoNCE across modality pairs \\
    Batch size & $256$ \\
    Epochs & $200$ \\
    Warmup & $10$ epochs \\
    Base learning rate & $1.5 \times 10^{-4}$ \\
    Weight decay & $0.05$ \\
    Schedule & Cosine learning-rate schedule \\
    \bottomrule
  \end{tabular}
\end{table}

\section{Additional Experimental Results and Controls}
\label{app:experiments}

\textbf{Control for contact-sequence overlap.}
Table~\ref{tab:app_overlap_control} verifies that the held-out-contact-sequence result is not due
to a weaker encoder. Evaluating the frame-random-trained encoder on the held-out-contact-sequence
pool gives essentially the same result as training directly under the held-out-contact-sequence
protocol.

\begin{table}[h]
  \centering
  \footnotesize
  \caption{Control for contact-sequence overlap. Both evaluations use the same
  held-out-contact-sequence test pool.}
  \label{tab:app_overlap_control}
  \begin{tabular}{lc}
    \toprule
    Encoder / evaluation & Tac$\rightarrow$text R@1 \\
    \midrule
    Same encoder as frame-random control, evaluated on held-out contact sequences & $62.33\%$ \\
    Encoder trained under held-out-contact-sequence protocol & $62.44\%$ \\
    \bottomrule
  \end{tabular}
\end{table}

\textbf{Frame-density controls for contact-sequence overlap.}
Table~\ref{tab:app_density_leakage} summarizes how much of the frame-random gain remains when
adjacent frames are removed at test time. Removing immediate neighbors reduces the effect, but a
substantial gap remains, suggesting that models can also exploit broader contact-sequence-specific
structure.

\begin{table}[h]
  \centering
  \footnotesize
  \caption{Frame-density controls for the contact-sequence-overlap mechanism.}
  \label{tab:app_density_leakage}
  \begin{tabular}{lcc}
    \toprule
    Test-time density & Description & Frame-random gain \\
    \midrule
    Full & All depths in the press & $+17.7$\,pp \\
    Uniform5 & Five frames from shallow to deep contact & $+12.3$\,pp \\
    Deep5 & Five deepest frames only & $+10.4$\,pp \\
    \bottomrule
  \end{tabular}
\end{table}

\textbf{Raw-pixel audit details for TVL/HCT.}
For each HCT test frame, we compute raw-pixel L2 distance to the nearest same-contact-sequence
training frame and to the nearest cross-sequence frame. In $350/356$ test frames ($98.3\%$), the
same-sequence neighbor is closer than any cross-sequence frame, with a median cross/same distance
ratio of $2.6\times$. This confirms that the released split can be exploited without learned
representations.

\textbf{Training-time frame sampling.}
Table~\ref{tab:app_recipe_full} reports the full training-density comparison including the
multi-positive tactile$\rightarrow$vision direction. Uniform5 improves over full-density training
while using one third of the tactile frames.

\begin{table}[h]
  \centering
  \footnotesize
  \caption{Effect of training-time frame sampling, evaluated on the same
  held-out-contact-sequence test pool.}
  \label{tab:app_recipe_full}
  \begin{tabular}{lccc}
    \toprule
    Training density & Training frames & Tac$\rightarrow$text R@1 & Tac$\rightarrow$vis R@1 \\
    \midrule
    Full press ($\sim$16 frames/sequence) & 22{,}576 & $62.44\%$ & $70.95\%$ \\
    Deep5 (five deepest frames)           & 7{,}055  & $59.52\%$ & $65.96\%$ \\
    Uniform5 (five frames across press)   & 7{,}055  & $\mathbf{68.18\%}$ & $\mathbf{78.29\%}$ \\
    \bottomrule
  \end{tabular}
\end{table}

\begin{figure}[h]
  \centering
  \includegraphics[width=0.75\linewidth]{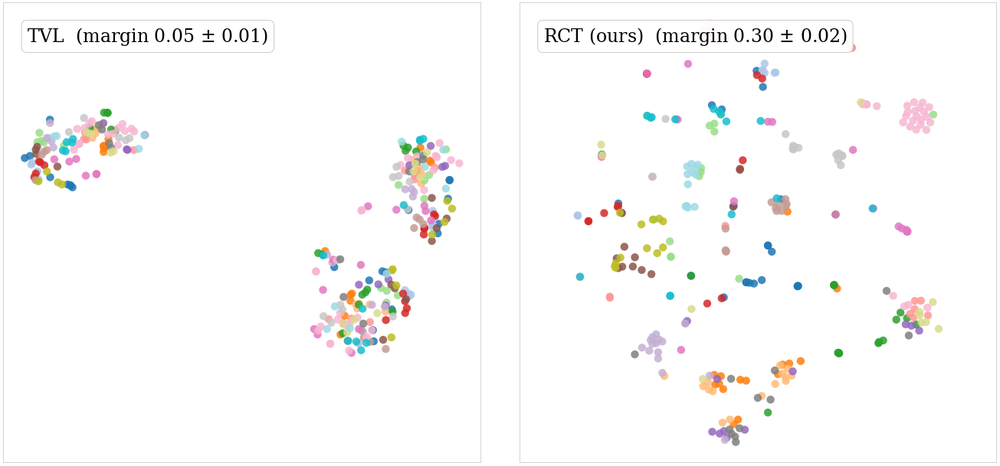}
  \caption{Qualitative visualization of tactile embeddings for $K=20$ held-out materials. The
  t-SNE plot illustrates the same trend quantified by the separability margin: RCT-trained
  embeddings form clearer material-level clusters than the released TVL encoder.}
  \label{fig:app_tsne}
\end{figure}

\textbf{Qualitative embedding visualization.}
Figure~\ref{fig:app_tsne} visualizes embeddings for $K=20$ held-out materials. The plot is
qualitative only; the quantitative evidence is the separability margin and downstream probe
performance in the main paper.

\textbf{Balanced and non-balanced downstream controls.}
We additionally check the downstream probe results under balanced and non-balanced choices of
held-out materials. The qualitative pattern is unchanged: RCT improves material-category probes,
whereas hard/soft prediction remains near the majority baseline. These controls support the
interpretation that category-level material information transfers better than binary hardness in
the current setup.

\textbf{Cross-domain transfer across training-data composition.}
Table~\ref{tab:app_transfer} reports per-dataset retrieval as a function of the training-data
composition, holding the architecture and all hyperparameters fixed. SSVTP and HCT are scored
with the diagonal tactile$\rightarrow$vision criterion, as in Table~\ref{tab:repro}, while RCT is
scored with tactile$\rightarrow$text R@1 on the $K=20$ material-disjoint hold-out, as in
Table~\ref{tab:novel}; RCT tactile$\rightarrow$vision is omitted here because the single
image-per-material design makes the diagonal criterion degenerate and requires the multi-positive
criterion. Two effects are visible. First, adding RCT to SSVTP or HCT does not degrade their
in-domain retrieval, indicating no negative transfer onto the existing benchmarks. Second,
adding SSVTP or HCT to RCT does not improve held-out-material generalization on RCT; the
HCT-containing mixes are in fact lower, because the importance sampler shifts most of the
training budget away from RCT. RCT-only training therefore remains the strongest setting for
held-out-material generalization on RCT.

\begin{table}[h]
  \centering
  \footnotesize
  \caption{Cross-domain transfer as a function of training-data composition. SSVTP and HCT report
  diagonal tactile$\rightarrow$vision R@1; RCT reports tactile$\rightarrow$text R@1 on the $K=20$
  material-disjoint hold-out. Diagonal in-domain cells are in bold.}
  \label{tab:app_transfer}
  \begin{tabular}{lccc}
    \toprule
    Training data & SSVTP Tac$\rightarrow$vis & HCT Tac$\rightarrow$vis & RCT ($K{=}20$) Tac$\rightarrow$text \\
    \midrule
    SSVTP only            & $\mathbf{40.22\%}$ & $0.56\%$           & $15.41\%$ \\
    HCT only              & $3.26\%$           & $\mathbf{87.50\%}$ & $12.16\%$ \\
    RCT only              & $2.17\%$           & $0.56\%$           & $\mathbf{20.35\%}$ \\
    SSVTP + RCT           & $42.39\%$          & $0.28\%$           & $20.18\%$ \\
    HCT + RCT             & $5.43\%$           & $87.08\%$          & $13.26\%$ \\
    SSVTP + HCT + RCT     & $36.96\%$          & $83.99\%$          & $13.41\%$ \\
    \bottomrule
  \end{tabular}
\end{table}

\textbf{Held-out position and held-out sensor.}
Table~\ref{tab:splits} lists held-out-position and held-out-sensor evaluation as supported
settings; Table~\ref{tab:app_axis} reports them. Both hold out a non-material axis while keeping
every material visible during training, so they isolate robustness to contact location and to
sensor instance rather than material-level generalization. Holding out a contact position is
easy, since the same material is still seen at four other positions. Holding out a DIGIT sensor
instance is as hard as holding out materials entirely, even though the materials themselves were
seen during training. This indicates that sensor-instance robustness is a distinct bottleneck:
the model does not absorb the physical tolerances and calibration differences between sensor
instances, and treats a held-out sensor much like an unseen material. Sensor-disjoint splits and
multiple sensor instances are therefore needed for honest evaluation. All rows use RCT-only
training at full frame density.

\begin{table}[h]
  \centering
  \footnotesize
  \caption{Held-out-position and held-out-sensor evaluation, compared with held-out-material.
  All settings use RCT-only training; tactile$\rightarrow$vision uses the multi-positive
  criterion.}
  \label{tab:app_axis}
  \begin{tabular}{lcc}
    \toprule
    Held-out axis (materials seen at train) & Tac$\rightarrow$text R@1 & Tac$\rightarrow$vis R@1 \\
    \midrule
    Contact position (p5; seen at p1--p4)        & $\mathbf{65.46\%}$ & $\mathbf{74.16\%}$ \\
    DIGIT sensor instance (d3; seen at d1, d2)   & $21.51\%$          & $13.00\%$ \\
    \midrule
    Held-out material ($K{=}20$; not seen)       & $20.35\%$          & $18.87\%$ \\
    \bottomrule
  \end{tabular}
\end{table}

\textbf{Seed robustness of held-out-material evaluation.}
A single held-out-material draw gives a noisy estimate, since the $K=20$ test pool is small and
its difficulty depends on which materials are sampled. Table~\ref{tab:app_seed} reports
tactile$\rightarrow$text R@1 across three seeds for two hold-out selection strategies: diverse
selection, which greedily maximizes descriptor coverage, and balanced selection, which balances
top-level categories across the hold-out. Diverse selection has high seed variance
($\pm 6.14$\,pp), so single-seed numbers should be read as draws from a wide distribution.
Balanced selection raises the mean and lowers the variance ($29.79 \pm 2.93$\,pp), since it
reduces the chance of a hold-out dominated by near-synonymous or unusually easy materials. We
recommend reporting the mean and standard deviation over seeds rather than a single draw.

\begin{table}[h]
  \centering
  \footnotesize
  \caption{Seed robustness of held-out-material evaluation. RCT tactile$\rightarrow$text R@1
  ($K=20$), RCT-only training, each encoder evaluated on its own seed-matched material-disjoint
  test pool.}
  \label{tab:app_seed}
  \begin{tabular}{lcccc}
    \toprule
    Hold-out selection & Seed 42 & Seed 0 & Seed 7 & Mean $\pm$ std \\
    \midrule
    Diverse (descriptor coverage)   & $20.35\%$ & $22.99\%$ & $32.06\%$ & $25.13 \pm 6.14$ \\
    Balanced (category-balanced)    & $32.24\%$ & $30.59\%$ & $26.54\%$ & $\mathbf{29.79 \pm 2.93}$ \\
    \bottomrule
  \end{tabular}
\end{table}

\section{Dataset Release}
\label{app:release}

\textbf{Material-ID correction.}
One material had an ID-format mismatch and was absent from the merged-annotator training labels, so diverse-selection models train on 121 of the 122 materials. This affects training coverage only, not the held-out test set. The released metadata already includes the correction.

\textbf{Release contents.}
The release includes tactile frames, material images, material-level descriptor annotations,
force signals, contact-sequence identifiers, material/category/sensor/position metadata, split
files for all held-out settings. Scripts for split generation and evaluation will be provided in the supplementary material. %

\end{document}